# Recurrent Residual Convolutional Neural Network based on U-Net (R2U-Net) for Medical Image Segmentation

Md Zahangir Alom[1*], *Student Member, IEEE*, Mahmudul Hasan[2], Chris Yakopcic[1], *Member, IEEE*,
Tarek M. Taha[1], *Member, IEEE*, and Vijayan K. Asari[1], *Senior Member, IEEE*

*Abstract*—Deep learning (DL) based semantic segmentation methods have been providing state-of-the-art performance in the last few years. More specifically, these techniques have been successfully applied to medical image classification, segmentation, and detection tasks. One deep learning technique, U-Net, has become one of the most popular for these applications. In this paper, we propose a Recurrent Convolutional Neural Network (RCNN) based on U-Net as well as a Recurrent Residual Convolutional Neural Network (RRCNN) based on U-Net models, which are named RU-Net and R2U-Net respectively. The proposed models utilize the power of U-Net, Residual Network, as well as RCNN. There are several advantages of these proposed architectures for segmentation tasks. First, a residual unit helps when training deep architecture. Second, feature accumulation with recurrent residual convolutional layers ensures better feature representation for segmentation tasks. Third, it allows us to design better U-Net architecture with same number of network parameters with better performance for medical image segmentation. The proposed models are tested on three benchmark datasets such as blood vessel segmentation in retina images, skin cancer segmentation, and lung lesion segmentation. The experimental results show superior performance on segmentation tasks compared to equivalent models including U-Net and residual U-Net (ResU-Net).

## I. INTRODUCTION

Nowadays DL provides state-of-the-art performance for image classification [1], segmentation [2], detection and tracking [3], and captioning [4]. Since 2012, several Deep Convolutional Neural Network (DCNN) models have been proposed such as AlexNet [1], VGG [5], GoogleNet [6], Residual Net [7], DenseNet [8], and CapsuleNet [9][65]. A DL based approach (CNN in particular) provides state-of-the-art performance for classification and segmentation tasks for several reasons: first, activation functions resolve training problems in DL approaches. Second, dropout helps regularize the networks. Third, several efficient optimization techniques are available for training CNN models [1]. However, in most cases, models are explored and evaluated using classification tasks on very large-scale datasets like ImageNet [1], where the outputs of the classification tasks are single label or probability values. Alternatively, small architecturally variant models are used for semantic image segmentation tasks. For example, a fully-connected convolutional neural network (FCN) also provides state-of-the-art results for image segmentation tasks in computer vision [2]. Another variant of FCN was also proposed which is called SegNet [10].

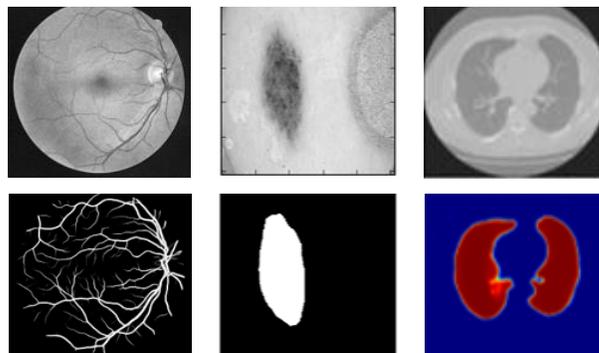

Fig. 1. Medical image segmentation: retina blood vessel segmentation in the left, skin cancer lesion segmentation, and lung segmentation in the right.

Due to the great success of DCNNs in the field of computer vision, different variants of this approach are applied in different modalities of medical imaging including segmentation, classification, detection, registration, and medical information processing. The medical imaging comes from different imaging techniques such as Computer Tomography (CT), ultrasound, X-ray, and Magnetic Resonance Imaging (MRI). The goal of Computer-Aided Diagnosis (CAD) is to obtain a faster and better diagnosis to ensure better treatment of a large number of people at the same time. Additionally, efficient automatic processing without human involvement to reduce human error and also reduces overall time and cost. Due to the slow process and tedious nature of

Md Zahangir Alom[1*], Chris Yakopcic[1], Tarek M. Taha[1], and Vijayan K. Asari[1] are with the University of Dayton, 300 College Park, Dayton, OH, 45469, USA. (e-mail: {alomm1, cyakopcic1, ttaha1, vasari1}@udayton.edu).

Mahmudul Hasan[2], is with Comcast Labs, Washington, DC, USA. (e-mail: mahmud.ucr@gmail.com).

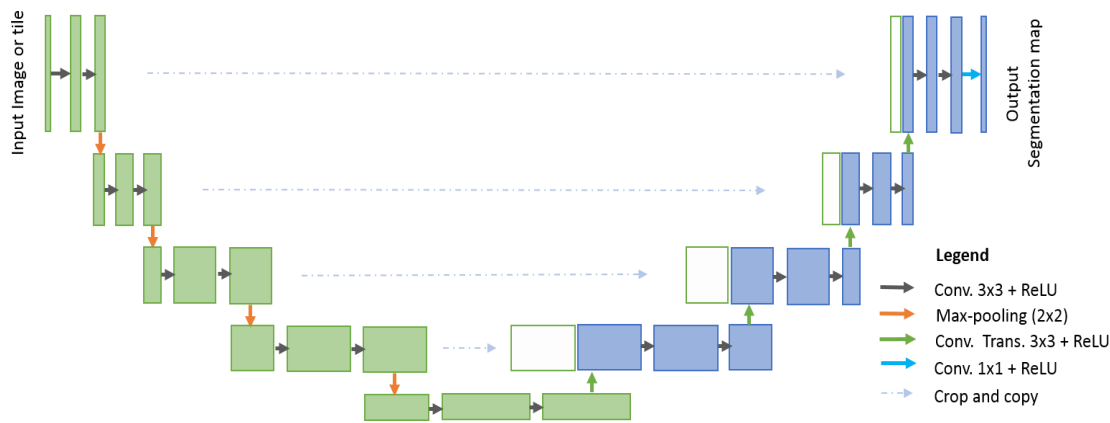

Fig. 2. U-Net architecture consisted with convolutional encoding and decoding units that take image as input and produce the segmentation feature maps with respective pixel classes.

manual segmentation approaches, there is a significant demand for computer algorithms that can do segmentation quickly and accurately without human interaction. However, there are some limitations of medical image segmentation including data scarcity and class imbalance. Most of the time the large number of labels (often in the thousands) for training is not available for several reasons [11]. Labeling the dataset requires an expert in this field which is expensive, and it requires a lot of effort and time. Sometimes, different data transformation or augmentation techniques (data whitening, rotation, translation, and scaling) are applied for increasing the number of labeled samples available [12, 13, and 14]. In addition, patch based approaches are used for solving class imbalance problems. In this work, we have evaluated the proposed approaches on both patch-based and entire image-based approaches. However, to switch from the patch-based approach to the pixel-based approach that works with the entire image, we must be aware of the class imbalance problem. In the case of semantic segmentation, the image backgrounds are assigned a label and the foreground regions are assigned a target class. Therefore, the class imbalance problem is resolved without any trouble. Two advanced techniques including cross-entropy loss and dice similarity are introduced for efficient training of classification and segmentation tasks in [13, 14].

Furthermore, in medical image processing, global localization and context modulation is very often applied for localization tasks. Each pixel is assigned a class label with a desired boundary that is related to the contour of the target lesion in identification tasks. To define these target lesion boundaries, we must emphasize the related pixels. Landmark detection in medical imaging [15, 16] is one example of this. There were several traditional machine learning and image processing techniques available for medical image segmentation tasks before the DL revolution, including amplitude segmentation based on histogram features [17], the region based segmentation method [18], and the graph-cut approach [19]. However, semantic segmentation approaches that utilize DL have become very popular in recent years in the field of medical image segmentation, lesion detection, and localization [20]. In addition, DL based approaches are known as universal learning approaches, where a single model can be utilized efficiently in different modalities of medical imaging such as MRI, CT, and X-ray.

According to a recent survey, DL approaches are applied to almost all modalities of medical imagining [20, 21]. Furthermore, the highest number of papers have been published on segmentation tasks in different modalities of medical imaging [20, 21]. A DCNN based brain tumor segmentation and detection method was proposed in [22].

From an architectural point of view, the CNN model for classification tasks requires an encoding unit and provides class probability as an output. In classification tasks, we have performed convolution operations with activation functions followed by sub-sampling layers which reduces the dimensionality of the feature maps. As the input samples traverse through the layers of the network, the number of feature maps increases but the dimensionality of the feature maps decreases. This is shown in the first part of the model (in green) in Fig. 2. Since, the number of feature maps increase in the deeper layers, the number of network parameters increases respectively. Eventually, the Softmax operations are applied at the end of the network to compute the probability of the target classes.

As opposed to classification tasks, the architecture of segmentation tasks requires both convolutional encoding and decoding units. The encoding unit is used to encode input images into a larger number of maps with lower dimensionality. The decoding unit is used to perform up-convolution (de-convolution) operations to produce segmentation maps with the same dimensionality as the original input image. Therefore, the architecture for segmentation tasks generally requires almost double the number of network parameters when compared to the architecture of the classification tasks. Thus, it is important to design efficient DCNN architectures for segmentation tasks which can ensure better performance with less number of network parameters.

This research demonstrates two modified and improved segmentation models, one using recurrent convolution networks, and another using recurrent residual convolutional networks. To accomplish our goals, the proposed models are evaluated on different modalities of medical imagining as shown in Fig. 1. The contributions of this work can be summarized as follows:

*1) Two new models RU-Net and R2U-Net are introduced for medical image segmentation.*
*2) The experiments are conducted on three different modalities of medical imaging including retina blood vessel segmentation, skin cancer segmentation, and lung segmentation.*
*3) Performance evaluation of the proposed models is conducted for the patch-based method for retina blood vessel segmentation tasks and the end-to-end image-based approach for skin lesion and lung segmentation tasks.*
*4) Comparison against recently proposed state-of-the-art methods that shows superior performance against equivalent models with same number of network parameters.*

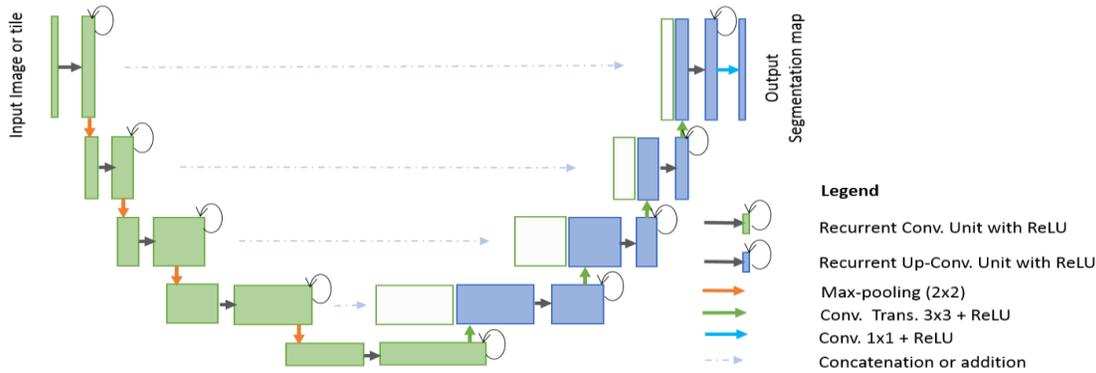

Fig. 3. RU-Net architecture with convolutional encoding and decoding units using recurrent convolutional layers (RCL) based U-Net architecture. The residual units are used with RCL for R2U-Net architecture.

The paper is organized as follows: Section II discusses related work. The architectures of the proposed RU-Net and R2U-Net models are presented in Section III. Section IV, explains the datasets, experiments, and results. The conclusion and future direction are discussed in Section V.

## II. RELATED WORK

Semantic segmentation is an active research area where DCNNs are used to classify each pixel in the image individually, which is fueled by different challenging datasets in the fields of computer vision and medical imaging [23, 24, and 25]. Before the deep learning revolution, the traditional machine learning approach mostly relied on hand engineered features that were used for classifying pixels independently. In the last few years, a lot of models have been proposed that have proved that deeper networks are better for recognition and segmentation tasks [5]. However, training very deep models is difficult due to the vanishing gradient problem, which is resolved by implementing modern activation functions such as Rectified Linear Units (ReLU) or Exponential Linear Units (ELU) [5,6]. Another solution to this problem is proposed by He et al., a deep residual model that overcomes the problem utilizing an identity mapping to facilitate the training process [26].

In addition, CNNs based segmentation methods based on FCN provide superior performance for natural image segmentation [2]. One of the image patch-based architectures is called Random architecture, which is very computationally intensive and contains around 134.5M network parameters. The main drawback of this approach is that a large number of pixel overlap and the same convolutions are performed many times. The performance of FCN has improved with recurrent neural networks (RNN), which are fine-tuned on very large datasets [27]. Semantic image segmentation with DeepLab is one of the state-of-the-art performing methods [28]. SegNet consists of two parts, one is the encoding network which is a 13-layer VGG16 network [5], and the corresponding decoding network uses pixel-wise classification layers. The main contribution of this paper is the way in which the decoder up-samples its lower resolution input feature maps [10]. Later, an improved version of SegNet, which is called Bayesian SegNet was proposed in 2015 [29]. Most of these architectures are explored using computer vision applications. However, there are some deep learning models that have been proposed specifically for the medical image segmentation, as they consider data insufficiency and class imbalance problems.

One of the very first and most popular approaches for semantic medical image segmentation is called "U-Net" [12]. A diagram of the basic U-Net model is shown in Fig. 2. According to the structure, the network consists of two main parts: the convolutional encoding and decoding units. The basic convolution operations are performed followed by ReLU activation in both parts of the network. For down sampling in the encoding unit, 2×2 max-pooling operations are performed. In the decoding phase, the convolution transpose (representing up-convolution, or de-convolution) operations are performed to up-sample the feature maps. The very first version of U-Net was used to crop and copy feature maps from the encoding unit to the decoding unit. The U-Net model provides several advantages for segmentation tasks: first, this model allows for the use of global location and context at the same time. Second, it works with very few training samples and provides better performance for segmentation tasks [12]. Third, an end-to-end pipeline process the entire image in the forward pass and directly produces segmentation maps. This ensures that U-Net preserves the full context of the input images, which is a major advantage when compared to patch-based segmentation approaches [12, 14].

However, U-Net is not only limited to the applications in the domain of medical imaging, nowadays this model is massively applied for computer vision tasks as well [30, 31]. Meanwhile, different variants of U-Net models have been proposed, including a very simple variant of U-Net for CNN-based segmentation of Medical Imaging data [32]. In this model, two modifications are made to the original design of U-Net: first, a combination of multiple segmentation maps and forward feature maps are summed (element-wise) from one part of the network to the other. The feature maps are taken from different layers of encoding and decoding units and finally summation (element-wise) is performed outside of the encoding and decoding units. The authors report promising performance improvement during training with better convergence compared to U-Net, but no benefit was observed when using a summation of features during the testing phase [32]. However, this concept proved that feature summation impacts the performance of a network. The importance of skipped connections for biomedical image segmentation tasks have been empirically evaluated with U-Net and residual networks [33]. A deep contour-aware network called Deep Contour-Aware Networks (DCAN) was proposed in 2016, which can extract multi-level contextual features using a hierarchical architecture for accurate gland segmentation of histology images and shows very good performance for segmentation [34]. Furthermore, Nabla-Net: a deep dig-like convolutional architecture was proposed for segmentation in 2017 [35].

Other deep learning approaches have been proposed based on U-Net for 3D medical image segmentation tasks as well. The 3D-Unet architecture for volumetric segmentation learns from sparsely annotated volumetric images [13]. A powerful end-to-end 3D medical image segmentation system based on volumetric images called V-net has been proposed, which consists of a FCN with residual connections [14]. This paper also introduces a dice loss layer [14]. Furthermore, a 3D deeply supervised approach for automated segmentation of volumetric medical images was presented in [36]. High-Res3DNet was proposed using residual networks for 3D segmentation tasks in 2016 [37]. In 2017, a CNN based brain tumor segmentation approach was proposed using a 3D-CNN model with a fully connected CRF [38]. Pancreas segmentation was proposed in [39], and Voxresnet was proposed in 2016 where a deep voxel wise residual network is used for brain segmentation. This architecture utilizes residual networks and summation of feature maps from different layers [40].

Alternatively, we have proposed two models for semantic segmentation based on the architecture of U-Net in this paper. The proposed Recurrent Convolutional Neural Networks (RCNN) model based on U-Net is named RU-Net, which is shown in Fig. 3. Additionally, we have proposed a residual RCNN based U-Net model which is called R2U-Net. The following section provides the architectural details of both models.

## III. RU-NET AND R2U-NET ARCHITECTURES

Inspired by the deep residual model [7], RCNN [41], and U-Net [12], we propose two models for segmentation tasks which are named RU-Net and R2U-Net. These two approaches utilize the strengths of all three recently developed deep learning models. RCNN and its variants have already shown superior performance on object recognition tasks using different benchmarks [42, 43]. The recurrent residual convolutional operations can be demonstrated mathematically according to the improved-residual networks in [43]. The operations of the Recurrent Convolutional Layers (RCL) are performed with respect to the discrete time steps that are expressed according to the RCNN [41]. Let's consider the $x_l$ input sample in the $l^{th}$ layer of the residual RCNN (RRCNN) block and a pixel located at $(i,j)$ in an input sample on the $k^{th}$ feature map in the RCL. Additionally, let's assume the output of the network $O_{ijk}^l(t)$ is at the time step $t$. The output can be expressed as follows as:

$$O_{ijk}^l(t) = \left(w_k^f\right)^T * x_l^{f(i,j)}(t) + (w_k^r)^T * x_l^{r(i,j)}(t-1) + b_k \qquad (1)$$

Here $x_l^{f(i,j)}(t)$ and $x_l^{r(i,j)}(t-1)$ are the inputs to the standard convolution layers and for the $l^{th}$ RCL respectively. The $w_k^f$ and $w_k^r$ values are the weights of the standard convolutional layer and the RCL of the $k^{th}$ feature map respectively, and $b_k$ is the bias. The outputs of RCL are fed to the standard ReLU activation function $f$ and are expressed:

$$\mathcal{F}(x_l, w_l) = f(O_{ijk}^l(t)) = \max(0, O_{ijk}^l(t)) \qquad (2)$$

$\mathcal{F}(x_l, w_l)$ represents the outputs from of $l^{th}$ layer of the RCNN unit. The output of $\mathcal{F}(x_l, w_l)$ is used for down-sampling and up-sampling layers in the convolutional encoding and decoding units of the RU-Net model respectively. In the case of R2U-Net, the final outputs of the RCNN unit are passed through the residual unit that is shown Fig. 4(d). Let's consider that the output of the RRCNN-block is $x_{l+1}$ and can be calculated as follows:

$$x_{l+1} = x_l + \mathcal{F}(x_l, w_l) \qquad (3)$$

Here, $x_l$ represents the input samples of the RRCNN-block. The $x_{l+1}$ sample is used the input for the immediate succeeding sub-sampling or up-sampling layers in the encoding and decoding convolutional units of R2U-Net. However, the number of feature maps and the dimensions of the feature maps for the residual units are the same as in the RRCNN-block shown in Fig. 4 (d).

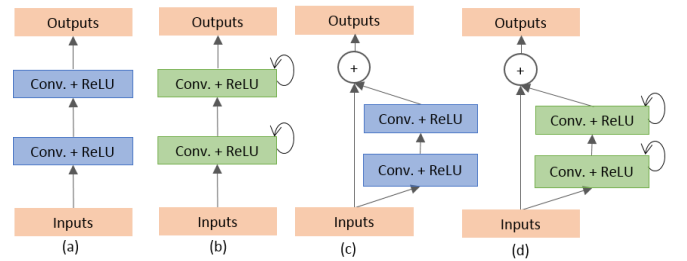

Fig. 4. Different variant of convolutional and recurrent convolutional units (a) Forward convolutional units, (b) Recurrent convolutional block (c) Residual convolutional unit, and (d) Recurrent Residual convolutional units (RRCU).

The proposed deep learning models are the building blocks of the stacked convolutional units shown in Fig. 4(b) and (d).

There are four different architectures evaluated in this work. First, U-Net with forward convolution layers and feature concatenation is applied as an alternative to the crop and copy method found in the primary version of U-Net [12]. The basic convolutional unit of this model is shown in Fig. 4(a). Second, U-Net with forward convolutional layers with residual connectivity is used, which is often called residual U-net (ResU-Net) and is shown in Fig. 4(c) [14]. The third architecture is U-Net with forward recurrent convolutional layers as shown in Fig. 4(b), which is named RU-Net. Finally, the last architecture is U-Net with recurrent convolution layers with residual connectivity as shown in Fig. 4(d), which is named R2U-Net. The pictorial representation of the unfolded RCL layers with respect to time-step is shown in Fig 5. Here $t=2$ (0 ~ 2), refers to the recurrent convolutional operation that includes one single convolution layer followed by two sub-sequential recurrent convolutional layers. In this implementation, we have applied concatenation to the feature maps from the encoding unit to the decoding unit for both RU-Net and R2U-Net models.

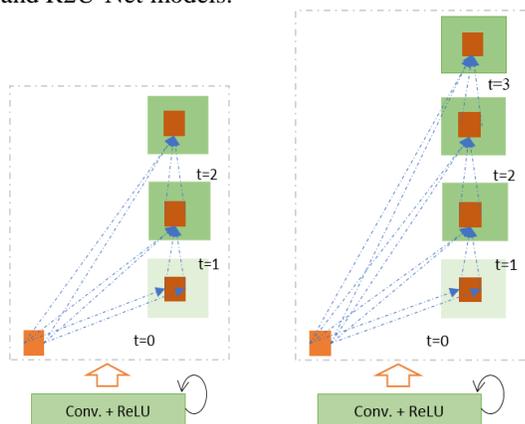

Fig. 5. Unfolded recurrent convolutional units for $t = 2$ (left) and $t = 3$ (right).

The differences between the proposed models with respect to the U-Net model are three-fold. This architecture consists of convolutional encoding and decoding units same as U-Net. However, the RCLs and RCLs with residual units are used instead of regular forward convolutional layers in both the encoding and decoding units. The residual unit with RCLs helps to develop a more efficient deeper model. Second, the efficient feature accumulation method is included in the RCL units of both proposed models. The effectiveness of feature accumulation from one part of the network to the other is shown in the CNN-based segmentation approach for medical imaging. In this model, the element-wise feature summation is performed outside of the U-Net model [32]. This model only shows the benefit during the training process in the form of better convergence. However, our proposed models show benefits for both training and testing phases due to the feature accumulation inside the model. The feature accumulation with respect to different time-steps ensures better and stronger feature representation. Thus, it helps extract very low-level features which are essential for segmentation tasks for different modalities of medical imaging (such as blood vessel segmentation). Third, we have removed the cropping and copying unit from the basic U-Net model and use only concatenation operations, resulting a much-sophisticated architecture that results in better performance.

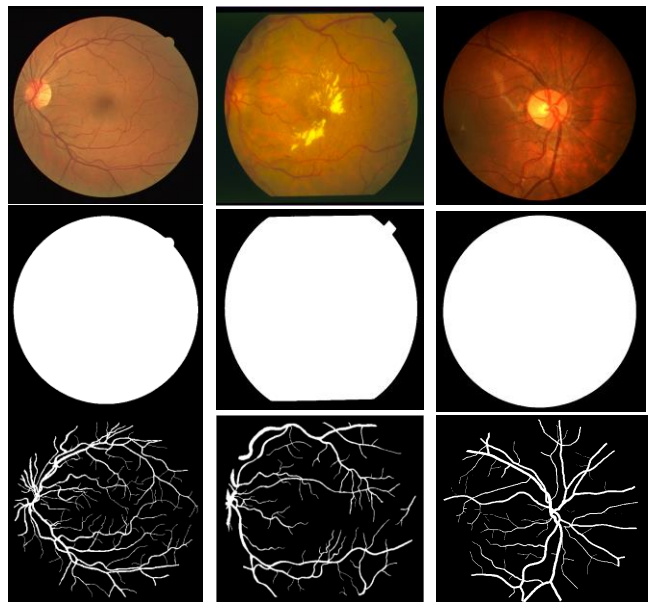

Fig. 6. Example images from training dataset: left column from DRIVE dataset, middle column from STARE dataset and right column from CHASE-DB1 dataset. The first row shows the original images, second row shows fields of view (FOV), and third row shows the target outputs.

There are several advantages of using the proposed architectures when compared with U-Net. The first is the efficiency in terms of the number of network parameters. The proposed RU-Net, and R2U-Net architectures are designed to have the same number of network parameters when compared to U-Net and ResU-Net, and RU-Net and R2U-Net show better performance on segmentation tasks. The recurrent and residual operations do not increase the number of network parameters. However, they do have a significant impact on training and testing performance. This is shown through empirical evidence with a set of experiments in the following sections [43]. This approach is also generalizable, as it easily be applied deep learning models based on SegNet [10], 3D-UNet [13], and V-Net [14] with improved performance for segmentation tasks.

## IV. EXPERIMENTAL SETUP AND RESULTS

To demonstrate the performance of the RU-Net and R2U-Net models, we have tested them on three different medical imaging datasets. These include blood vessel segmentations from retina images (DRIVE, STARE, and CHASE_DB1 shown in Fig. 6), skin cancer lesion segmentation, and lung segmentation from 2D images. For this implementation, the Keras, and TensorFlow frameworks are used on a single GPU machine with 56G of RAM and an NIVIDIA GEFORCE GTX-980 Ti.

### A. Database Summary
### 1) Blood Vessel Segmentation

We have experimented on three different popular datasets for retina blood vessel segmentation including DRIVE, STARE, and CHASH_DB1. The DRIVE dataset is consisted of 40 color retinal images in total, in which 20 samples are used for training and remaining 20 samples are used for testing. The size of each original image is 565×584 pixels [44]. To develop a square dataset, the images are cropped to only contain the data from columns 9 through 574, which then makes each image 565×565 pixels. In this implementation, we considered 190,000 randomly selected patches from 20 of the images in the DRIVE dataset, where 171,000 patches are used for training, and the remaining 19,000 patches used for validation. The size of each patch is 48×48 for all three datasets shown in Fig. 7. The second dataset, STARE, contains 20 color images, and each image has a size of 700×605 pixels [45, 46]. Due to the smaller number of samples, two approaches are applied very often for training and testing on this dataset. First, training sometimes performed with randomly selected samples from all 20 images [53].

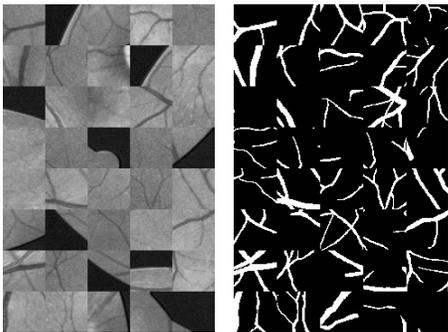

Fig. 7. Example patches in the left and corresponding outputs of patches are shown in the right.

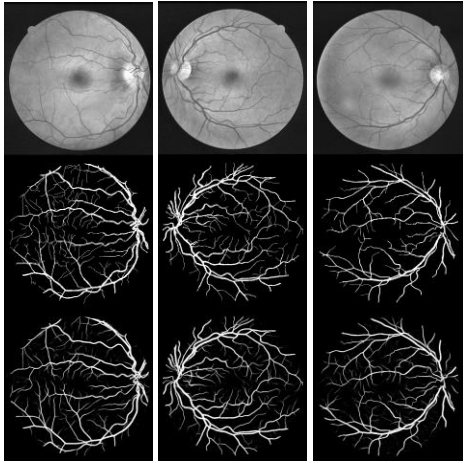

Fig. 8. Experimental outputs for DRIVE dataset using R2UNet: first row shows input image in gray scale, second row show ground truth, and third row shows the experimental outputs.

Another approach is the "leave-one-out" method, in which each image is tested, and training is conducted on the remaining 19 samples [47]. Therefore, there is no overlap between training and testing samples. In this implementation, we used the "leave-one-out" approach for STARE dataset. The CHASH_DB1 dataset contains 28 color retina images and the size of each image is 999×960 pixels [48]. The images in this dataset were collected from both left and right eyes of 14 school children. The dataset is divided into two sets where samples are selected randomly. A 20-sample set is used for training and the remaining 8 samples are used for testing.

As the dimensionality of the input data larger than the entire DRIVE dataset, we have considered 250,000 patches in total from 20 images for both STARE and CHASE_DB1. In this case 225,000 patches are used for training and the remaining 25,000 patches are used for validation. Since the binary FOV (which is shown in second row in Fig. 6) is not available for the STARE and CHASE_DB1 datasets, we generated FOV masks using a similar technique to the one described in [47]. One advantage of the patch-based approach is that the patches give the network access to local information about the pixels, which has impact on overall prediction. Furthermore, it ensures that the classes of the input data are balanced. The input patches are randomly sampled over an entire image, which also includes the outside region of the FOV.

*2) Skin Cancer Segmentation*

This dataset is taken from the Kaggle competition on skin lesion segmentation that occurred in 2017 [49]. This dataset contains 2000 samples in total. It consists of 1250 training samples, 150 validation samples, and 600 testing samples. The original size of each sample was 700×900, which was rescaled to 256×256 for this implementation. The training samples include the original images, as well as corresponding target binary images containing cancer or non-cancer lesions. The target pixels are represented with a value of either 255 or 0 for the pixels outside of the target lesion.

*3) Lung Segmentation*

The Lung Nodule Analysis (LUNA) competition at the Kaggle Data Science Bowl in 2017 was held to find lung lesions in 2D and 3D CT images. The provided dataset consisted of 534 2D samples with respective label images for lung segmentation [50]. For this study, 70% of the images are used for training and the remaining 30% are used for testing. The original image size was 512×512, however, we resized the images to 256×256 pixels in this implementation.

B. *Quantitative Analysis Approaches*

For quantitative analysis of the experimental results, several performance metrics are considered, including accuracy (AC), sensitivity (SE), specificity (SP), F1-score, Dice coefficient (DC), and Jaccard similarity (JS). To do this we also use the variables True Positive (TP), True Negative (TN), False Positive (FP), and False Negative (FN). The overall accuracy is calculated using Eq. (4), and sensitivity is calculated using Eq. (5).

$$AC = \frac{TP+TN}{TP+TN+FP+FN} \quad (4)$$

$$SE = \frac{TP}{TP+FN} \quad (5)$$

Furthermore, specificity is calculated using the following Eq. (6).

$$SP = \frac{TN}{TN+FP} \quad (6)$$

The DC is expressed as in Eq. (7) according to [51]. Here GT refers to the ground truth and SR refers the segmentation result.

$$DC = 2\frac{|GT \cap SR|}{|GT|+|SR|} \quad (7)$$

The JS is represented using Eq. (8) as in [52].

$$JS = \frac{|GT \cap SR|}{|GT \cup SR|} \quad (8)$$

However, the area under curve (AUC) and the receiver operating characteristics (ROC) curve are common evaluation measures for medical image segmentation tasks. In this experiment, we utilized both analytical methods to evaluate the performance of the proposed approaches considering the mentioned criterions against existing state-of-the-art techniques.

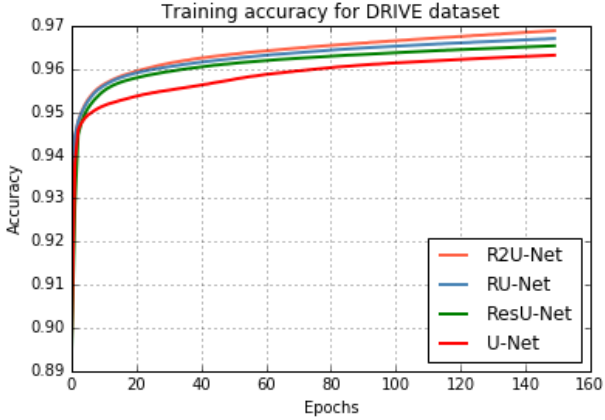

Fig. 9. Training accuracy of the proposed models of RU-Net, and R2U-Net against ResU-Net and U-Net.

### C. Results

#### 1) Retina Blood Vessel Segmentation Using the DRIVE Dataset

The precise segmentation results achieved with the proposed R2U-Net model are shown in Fig. 8. Figs. 9 and 10 show the training and validation accuracy when using the DRIVE dataset. These figures show that the proposed R2U-Net and RU-Net models provide better performance during both the training and validation phase when compared to U-Net and ResU-Net.

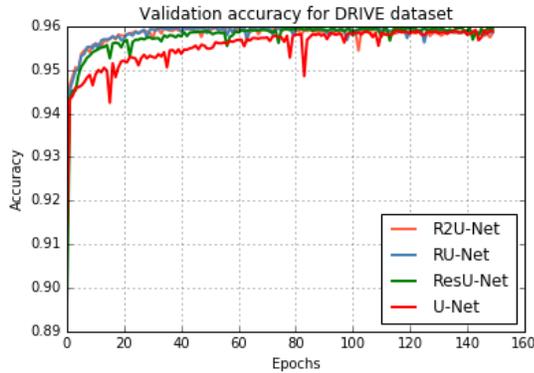

Fig. 10. Validation accuracy of the proposed models against ResU-Net and U-Net.

#### 2) Retina blood vessel segmentation on the STARE dataset

The experimental outputs of R2U-Net when using the STARE dataset are shown in Fig. 11. The training and validation accuracy for the STARE dataset is shown in Figs. 12 and 13 respectively.

R2U-Net shows a better performance than all other models during training. In addition, the validation accuracy in Fig. 13 demonstrates that the RU-Net and R2U-Net models provide better validation accuracy when compared to the equivalent U-Net and ResU-Net models. Thus, the performance demonstrates the effectiveness of the proposed approaches for segmentation tasks.

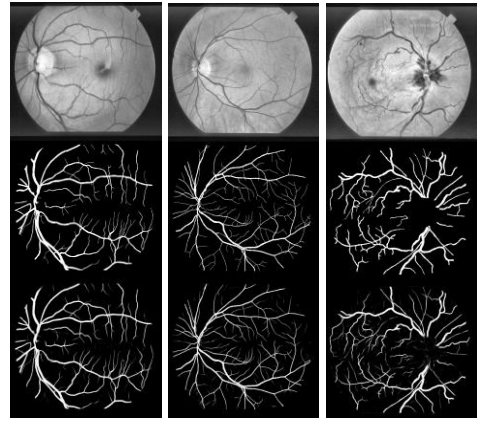

Fig. 11. Experimental outputs of STARE dataset using R2UNet: first row shows input image after performing normalization, second row show ground truth, and third row shows the experimental outputs.

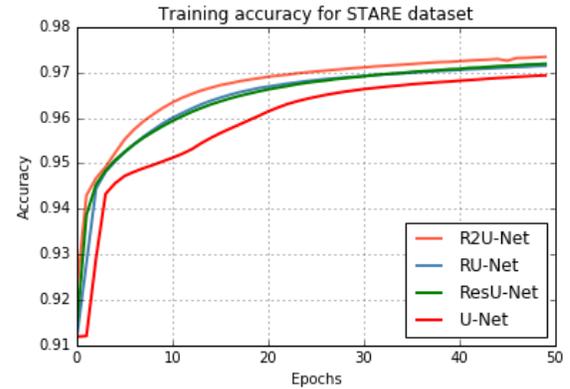

Fig. 12. Training accuracy in STARE dataset for R2U-Net, RU-Net, ResU-Net, and U-Net.

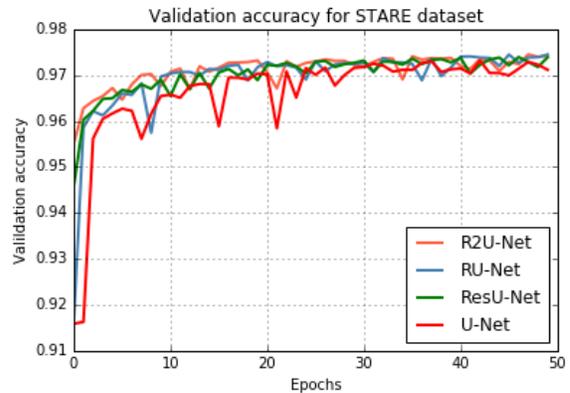

Fig. 13. Validation accuracy in STARE dataset for R2U-Net, RU-Net, ResU-Net, and U-Net.

#### 3) CHASE_DB1

For qualitative analysis, the example outputs of R2U-Net are shown in Fig. 14. For quantitative analysis, the results are given in Table I. From the table, it can be concluded that in all cases, the proposed RU-Net and R2U-Net models show better performance in terms of AUC and accuracy. The ROC for the highest AUCs for the R2U-Net model on each of the three retina blood vessel segmentation datasets is shown in Fig. 15.

validation during training with a batch size of 32 and 150 epochs.

The training accuracy of the proposed models R2U-Net and RU-Net was compared with that of ResU-Net and U-Net for an end-to-end image based segmentation approach. The result is

TABLE I. EXPERIMENTAL RESULTS OF PROPOSED APPROACHES FOR RETINA BLOOD VESSEL SEGMENTATION AND COMPARISON AGAINST OTHER TRADITIONAL AND DEEP LEARNING-BASED APPROACHES.

| Dataset | Methods | Year | F1-score | SE | SP | AC | AUC |
|---|---|---|---|---|---|---|---|
| DRIVE | Chen [53] | 2014 | - | o.7252 | 0.9798 | 0.9474 | 0.9648 |
| | Azzopardi [54] | 2015 | - | 0.7655 | 0.9704 | 0.9442 | 0.9614 |
| | Roychowdhury[55] | 2016 | - | 0.7250 | 0.9830 | 0.9520 | 0.9620 |
| | Liskowsk [56] | 2016 | - | 0.7763 | 0.9768 | 0.9495 | 0.9720 |
| | Qiaoliang Li [57] | 2016 | - | 0.7569 | 0.9816 | 0.9527 | 0.9738 |
| | U-Net | 2018 | 0.8142 | 0.7537 | 0.9820 | 0.9531 | 0.9755 |
| | Residual U-Net | 2018 | 0.8149 | 0.7726 | 0.9820 | 0.9553 | 0.9779 |
| | **Recurrent U-Net** | **2018** | **0.8155** | **0.7751** | **0.9816** | **0.9556** | **0.9782** |
| | **R2U-Net** | **2018** | **0.8171** | **0.7792** | **0.9813** | **0.9556** | **0.9784** |
| STARE | Marin et al. [58] | 2011 | - | 0.6940 | 0.9770 | 0.9520 | 0.9820 |
| | Fraz [59] | 2012 | - | 0.7548 | 0.9763 | 0.9534 | 0.9768 |
| | Roychowdhury[55] | 2016 | - | 0.7720 | 0.9730 | 0.9510 | 0.9690 |
| | Liskowsk [56] | 2016 | - | 0.7867 | 0.9754 | 0.9566 | 0.9785 |
| | Qiaoliang Li [57] | 2016 | - | 0.7726 | 0.9844 | 0.9628 | 0.9879 |
| | U-Net | 2018 | 0.8373 | 0.8270 | 0.9842 | 0.9690 | 0.9898 |
| | Residual U-Net | 2018 | 0.8388 | 0.8203 | 0.9856 | 0.9700 | 0.9904 |
| | **Recurrent U-Net** | **2018** | **0.8396** | **0.8108** | **0.9871** | **0.9706** | **0.9909** |
| | **R2U-Net** | **2018** | **0.8475** | **0.8298** | **0.9862** | **0.9712** | **0.9914** |
| CHASE_DB1 | Fraz [59] | 2012 | - | 0.7224 | 0.9711 | 0.9469 | 0.9712 |
| | Fraz [60] | 2014 | - | - | - | 0.9524 | 0.9760 |
| | Azzopardi [54] | 2015 | - | 0.7655 | 0.9704 | 0.9442 | 0.9614 |
| | Roychowdhury[55] | 2016 | - | 0.7201 | 0.9824 | 0.9530 | 0.9532 |
| | Qiaoliang Li [57] | 2016 | - | 0.7507 | 0.9793 | 0.9581 | 0.9793 |
| | U-Net | 2018 | 0.7783 | 0.8288 | 0.9701 | 0.9578 | 0.9772 |
| | Residual U-Net | 2018 | 0.7800 | 0.7726 | 0.9820 | 0.9553 | 0.9779 |
| | **Recurrent U-Net** | **2018** | **0.7810** | **0.7459** | **0.9836** | **0.9622** | **0.9803** |
| | **R2U-Net** | **2018** | **0.7928** | **0.7756** | **0.9820** | **0.9634** | **0.9815** |

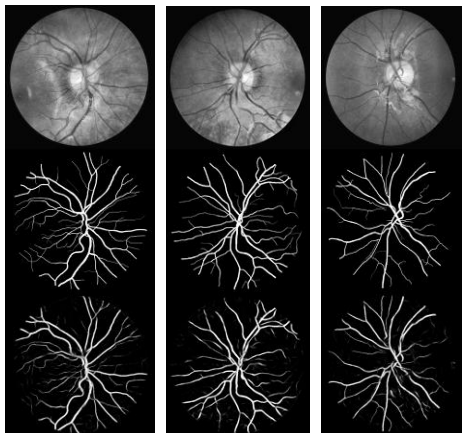

Fig. 14. Qualitative analysis for CHASE_DB1 dataset. The segmentation outputs of 8 testing samples using R2U-Net. First row shows the input images, second row is ground truth, and third row shows the segmentation outputs using R2U-Net.

*4) Skin Cancer Lesion Segmentation*

In this implementation, this dataset is preprocessed with mean subtraction and normalized according to the standard deviation. We used the ADAM optimization technique with a learning rate of $2\times10^{-4}$ and binary cross entropy loss. In addition, we also calculated MSE error during the training and validation phase. In this case 10% of the samples are used for shown in Fig. 16. The validation accuracy is shown in Fig. 17. In both cases, the proposed models show better performance when compared with the equivalent U-Net and ResU-Net models. This clearly demonstrates the robustness of the proposed models in end-to-end image-based segmentation tasks.

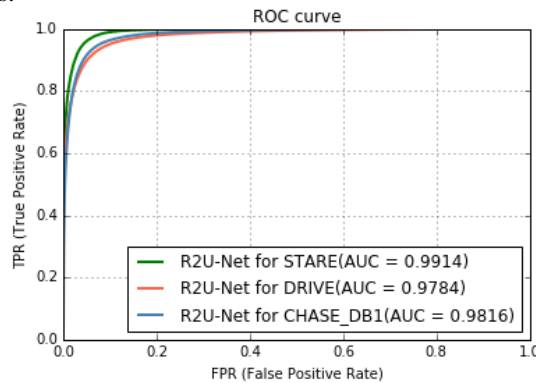

Fig. 15. AUC for retina blood vessel segmentation for the best performance achieved with R2U-Net.

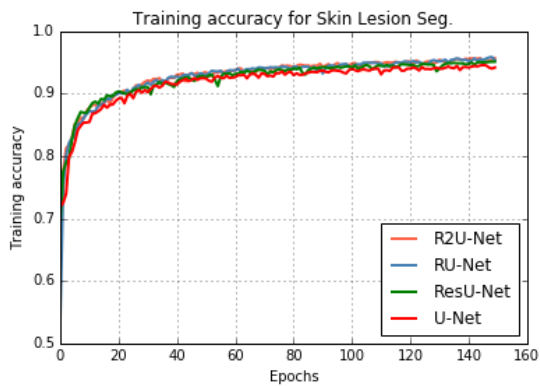

Fig. 16. Training accuracy for skin lesion segmentation.

clearly shows the robustness of the proposed segmentation method.

We have compared the performance of the proposed approaches against recently published results with respect to sensitivity, specificity, accuracy, AUC, and DC. The proposed R2U-Net model provides a testing accuracy 0.9424 with a higher AUC, which is 0.9419. The average AUC for skin lesion segmentation is shown in Fig. 19. In addition, we calculated the average DC in the testing phase and achieved 0.8616, which is around 1.26% better than recently proposed alternatives [62]. Furthermore, the JSC and F1 scores are calculated and the R2U-Net model obtains 0.9421 for JSC and 0.8920 for F1 score for skin lesion segmentation with $t=3$. These results are achieved

TABLE II. EXPERIMENTAL RESULTS OF PROPOSED APPROACHES FOR SKIN CANCER LESION SEGMENTATION AND COMPARISON AGAINST OTHER EXISTING APPROACHES. JACCARD SIMILARITY SCORE (JSC).

| Methods | Year | SE | SP | JSC | F1-score | AC | AUC | DC |
|---|---|---|---|---|---|---|---|---|
| Conv. classifier VGG-16 [61] | 2017 | 0.533 | - | - | - | 0.6130 | 0.6420 | - |
| Conv. classifier Inception-v3[61] | 2017 | 0.760 | - | - | - | 0.6930 | 0.7390 | - |
| Melanoma detection [62] | 2017 | - | - | - | - | o.9340 | - | 0.8490 |
| Skin Lesion Analysis [63] | 2017 | 0.8250 | 0.9750 | - | - | 0.9340 | - | - |
| U-Net (t=2) | 2018 | 0.9479 | 0.9263 | 0.9314 | 0.8682 | 0.9314 | 0.9371 | 0.8476 |
| ResU-Net (t=2) | 2018 | 0.9454 | 0.9338 | 0.9367 | 0.8799 | 0.9367 | 0.9396 | 0.8567 |
| RecU-Net (t=2) | 2018 | 0.9334 | 0.9395 | 0.9380 | 0.8841 | 0.9380 | 0.9364 | 0.8592 |
| **R2U-Net (t=2)** | 2018 | **0.9496** | 0.9313 | 0.9372 | 0.8823 | **0.9372** | 0.9405 | **0.8608** |
| **R2U-Net (t=3)** | 2018 | **0.9414** | 0.9425 | 0.9421 | 0.8920 | **0.9424** | 0.9419 | **0.8616** |

The quantitative results of this experiment were compared against existing methods as shown in Table II. Some of the example outputs from the testing phase are shown in Fig. 18. The first column shows the input images, the second column shows the ground truth, the network outputs are shown in the third column, and the fourth column demonstrates the final outputs after performing post processing with a threshold of 0.5. Figure 18 shows promising segmentation results.

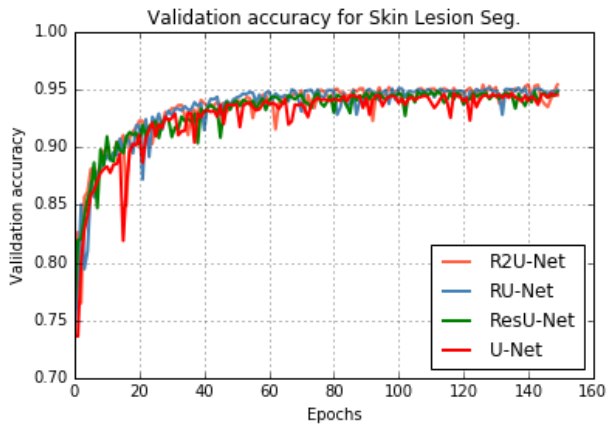

Fig. 17. Validation accuracy for skin lesion segmentation.

In most cases, the target lesions are segmented accurately with almost the same shape of ground truth. However, if we observe the second and third rows in Fig. 18, it can be clearly seen that the input images contain two spots, one is a target lesion and the other bright spot which is not a target. This result is obtained even though the non-target lesion is brighter than the target lesion shown in the third row in Fig. 18. The R2U-Net model still segments the desired part accurately, which with a R2U-Net model that only contains about 1.037 million (M) network parameters. Contrarily, the work presented in [61] evaluated VGG-16 and Incpetion-V3 models for skin lesion segmentation, but those networks contained around 138M and 23M network parameters respectively.

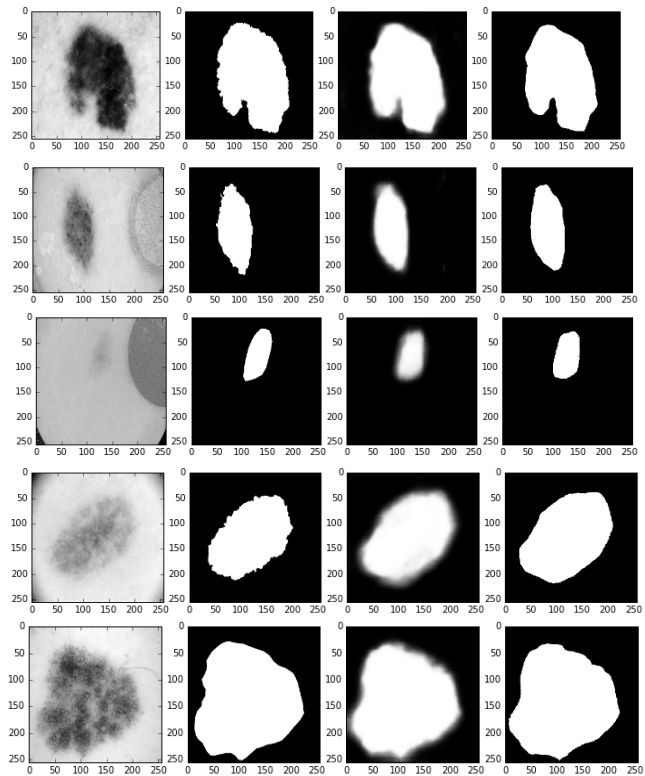

Fig. 18. This results demonstrates qualitative assessment of the proposed R2U-Net for skin cancer segmentation task with $t=3$. First column is the input sample, second column is ground truth, third column shows the outputs from network, and fourth column show the final resulting after performing thresholding with 0.5.

*5) Lung Segmentation*

Lung segmentation is very important for analyzing lung related diseases, and can be applied to lung cancer segmentation and lung pattern classification for identifying other problems. In this experiment, the ADAM optimizer is used with a learning rate of $2\times10^{-4}$. We used binary cross entropy loss, and also calculated MSE during training and validation. In this case 10% of the samples were used for validation with a batch size of 16 and 150 epochs 150. Table III shows the summary of how well the proposed models performed against equivalent U-Net and ResU-Net models. The experimental results show that the proposed models outperform the U-Net and ResU-Net models respectively. However, we also experimented with U-Net, ResU-Net, RU-Net, and R2U-Net models with following structure: 1→16→32→64→128→64 → 32→16→1. In this case we used a time-step of $t=3$, which refers to one forward convolution layer followed by three subsequent recurrent convolutional layers. This network was tested on skin and lung lesion segmentation. Though the number of network parameters increase little bit with respect to the time-step in the recurrent convolution layer, further improved performance can be clearly seen in the last rows of Table II and III. Furthermore, we have evaluated both of the proposed models for patch-based modeling on retina blood vessel segmentation and end-to-end image-based methods for skin and lung lesion segmentation.

In both cases, the proposed models outperform existing state-with same number of network parameters.

TABLE III. EXPERIMENTAL OUTPUTS OF PROPOSED MODELS OF RU-NET AND R2U-NET FOR LUNG SEGMENTATION AND COMPARISON AGAINST RESU-NET AND U-NET MODELS.

| Methods | Year | SE | SP | JSC | F1-Score | AC | AUC |
|---|---|---|---|---|---|---|---|
| U-Net (t=2) | 2018 | 0.9696 | 0.9872 | 0.9858 | 0.9658 | 0.9828 | 0.9784 |
| ResU-Net(t=2) | 2018 | 0.9555 | 0.9945 | 0.9850 | 0.9690 | 0.9849 | 0.9750 |
| RU-Net (t=2) | 2018 | 0.9734 | 0.9866 | 0.9836 | 0.9638 | 0.9836 | 0.9800 |
| R2U-Net (t=2) | 2018 | 0.9826 | 0.9918 | 0.9897 | 0.9780 | 0.9897 | 0.9872 |
| **R2U-Net (t=3)** | **2018** | **0.9832** | **0.9944** | **0.9918** | **0.9823** | **0.9918** | **0.9889** |

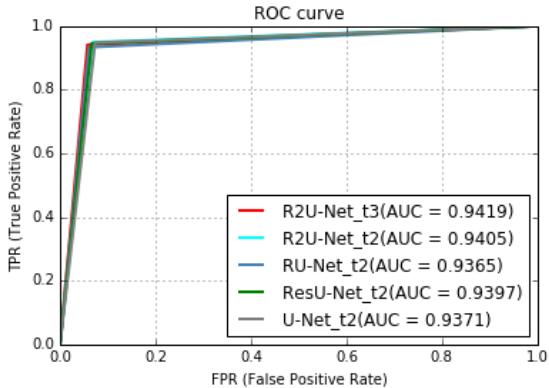

Fig. 19. ROC-AUC for skin segmentation four models with $t=2$ and $t=3$.

Furthermore, many models struggle to define the class boundary properly during segmentation tasks [64]. However, if we observe the experimental outputs shown in Fig. 20, the outputs in the third column show different hit maps on the border, which can be used to define the boundary of the lung region, while the ground truth tends to have a smooth boundary.

In addition, if we observe the input, ground truth, and output of this proposed approaches in the second row, it can be observed that the output of the proposed approaches shows better segmentation with appropriate contour. The ROC with AUCs are shown Fig. 21. The highest AUC is achieved with the proposed approach of R2U-Net with $t=3$.

*D. Evaluation*

Most of the cases, the networks are evaluated for different segmentation tasks with following architectures: 1→64→128→256→512→256 → 128→64→1 that require 4.2M network parameters and 1→64→128→256→512→256 → 128→64→1, which require about 8.5M network parameters

of-the-art methods including ResU-Net and U-Net in terms of AUC and accuracy on all three datasets. The network architectures with different numbers of network parameters with respect to the different time-step are shown in Table IV. The processing times during the testing phase for the STARE, CHASE_DB, and DRIVE datasets were 6.42, 8.66, and 2.84 seconds per sample respectively. In addition, skin cancer segmentation and lung segmentation take 0.22 and 1.145 seconds per sample respectively.

TABLE IV. ARCHITECTURE AND NUMBER OF NETWORK PARAMETERS.

| t | Network architectures | Number of parameters (million) |
|---|---|---|
| 2 | 1-> 16->32->64>128->64 –> 32->16->1 | 0.845 |
| 3 | 1-> 16->32->64->128->64 –> 32->16->1 | 1.037 |

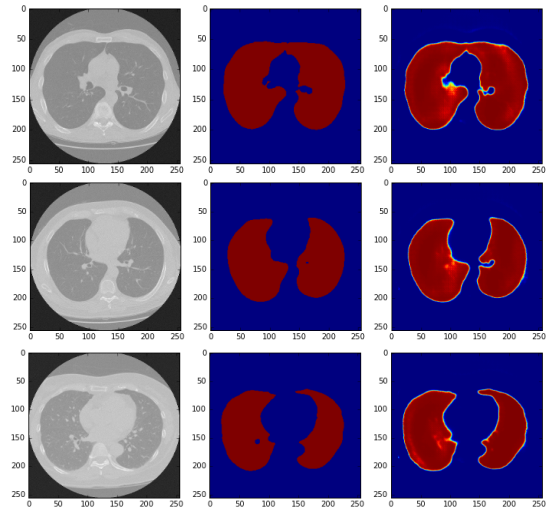

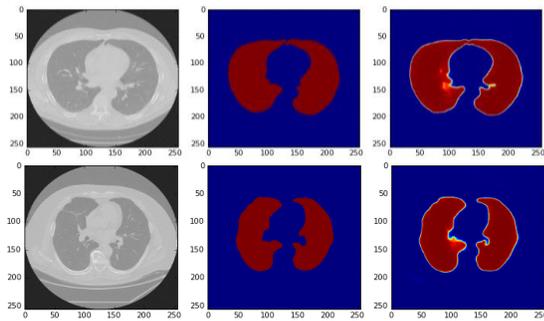

Fig. 20. Qualitative assessment of R2U-Net performance on Lung segmentation dataset: first column input images, second column ground truth, and third column outputs with R2U-Net.

*E. Computational time*

The computational time for testing per sample is shown in Table V for blood vessel segmentation for retina images, skin cancer, and lung segmentation respectively.

TABLE V. COMPUTATIONAL TIME FOR TESTING PHASE.

| Dataset | | Time (Sec.)/ sample |
|---|---|---|
| Blood vessel segmentation | DRIVE | 6.42 |
| | STARE | 8.66 |
| | CHASE_DB1 | 2.84 |
| Skin cancer segmentation | | 0.22 |
| Lung segmentation | | 1.15 |

## V. CONCLUSION AND FUTURE WORKS

In this paper, we proposed an extension of the U-Net architecture using Recurrent Convolutional Neural Networks and Recurrent Residual Convolutional Neural Networks. The proposed models are called "RU-Net" and "R2U-Net" respectively. These models were evaluated using three different applications in the field of medical imaging including retina blood vessel segmentation, skin cancer lesion segmentation, and lung segmentation. The experimental results demonstrate that the proposed RU-Net, and R2U-Net models show better performance in segmentation tasks with the same number of network parameters when compared to existing methods including the U-Net and residual U-Net (or ResU-Net) models on all three datasets. In addition, results show that these proposed models not only ensure better performance during the training but also in testing phase. In future, we would like to explore the same architecture with a novel feature fusion strategy from encoding to the decoding units.

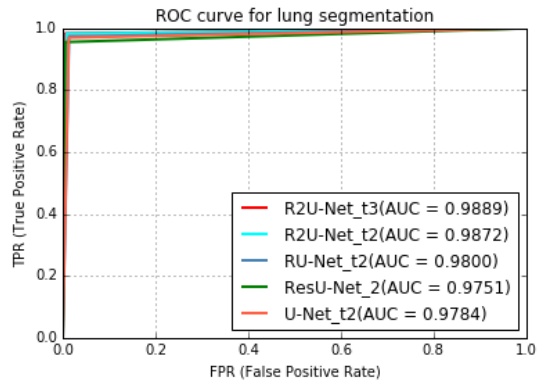

Fig. 21. ROC curve for lung segmentation four models with *t=2* and *t=3*.